\documentclass{article}

\usepackage{microtype}
\usepackage{graphicx}
\usepackage{subfigure}
\usepackage{booktabs} 
\usepackage{diagbox}
\usepackage{multirow}
\usepackage{hyperref}

\usepackage[accepted]{icml2025}

\usepackage{amsmath}
\usepackage{amssymb}
\usepackage{mathtools}
\usepackage{amsthm}

\usepackage[capitalize,noabbrev]{cleveref}

\theoremstyle{plain}

\theoremstyle{definition}

\theoremstyle{remark}

\usepackage[textsize=tiny]{todonotes}

\icmltitlerunning{Kolmogorov–Arnold Neural Interpolator for Downscaling and Correcting Meteorological Fields from In-Situ Observations}

\begin{document}

\twocolumn[
\icmltitle{Kolmogorov–Arnold Neural Interpolator for Downscaling and Correcting Meteorological Fields from In-Situ Observations}




\begin{icmlauthorlist}
\icmlauthor{Zili Liu}{sch,lab}
\icmlauthor{Hao Chen}{lab}
\icmlauthor{Lei Bai}{lab}
\icmlauthor{Wenyuan Li}{hku}
\icmlauthor{Wanli Ouyang}{lab}
\icmlauthor{Zhengxia Zou}{gnc}
\icmlauthor{Zhenwei Shi}{sch}

\end{icmlauthorlist}

\icmlaffiliation{sch}{Image Processing Center, School of Astronautics, Beihang University, Beijing, China}
\icmlaffiliation{lab}{Shanghai Artificial Intelligence Laboratory, Shanghai, China}
\icmlaffiliation{hku}{Department of Geography, University of Hong Kong, Hong Kong, China}
\icmlaffiliation{gnc}{Department of Guidance, Navigation and Control, School of Astronautics, Beihang University, Beijing 100191, China}

\icmlcorrespondingauthor{Zhenwei Shi}{shizhenwei@buaa.edu.cn}
\icmlcorrespondingauthor{Hao Chen}{chenhao1@pjlab.org.cn}

\icmlkeywords{Machine Learning, ICML}

\vskip 0.3in
]



\printAffiliationsAndNotice{}  

\begin{abstract}

Obtaining accurate weather forecasts at station locations is a critical challenge due to systematic biases arising from the mismatch between multi-scale, continuous atmospheric characteristic and their discrete, gridded representations. Previous works have primarily focused on modeling gridded meteorological data, inherently neglecting the off-grid, continuous nature of atmospheric states and leaving such biases unresolved. To address this, we propose the \emph{Kolmogorov–Arnold Neural Interpolator} (KANI), a novel framework that redefines meteorological field representation as continuous neural functions derived from discretized grids. Grounded in the Kolmogorov–Arnold theorem, KANI captures the inherent continuity of atmospheric states and leverages sparse in-situ observations to correct these biases systematically. Furthermore, KANI introduces an innovative \emph{zero-shot} downscaling capability, guided by high-resolution topographic textures without requiring high-resolution meteorological fields for supervision. Experimental results across three sub-regions of the continental United States indicate that KANI achieves an accuracy improvement of 40.28\% for temperature and 67.41\% for wind speed, highlighting its significant improvement over traditional interpolation methods. This enables continuous neural representation of meteorological variables through neural networks, transcending the limitations of conventional grid-based representations.

\end{abstract}

\vspace{-5pt}
\section{Introduction}
\label{intro}
Deep learning-based weather forecasting has achieved remarkable progress in recent years \cite{bi2023accurate,chen2023fengwu,han2024fengwu}, demonstrating superior performance across various tasks including data simulation \cite{pmlr-v235-xiao24a,pmlr-v235-huang24h,xiang2024adaf} and precipitation nowcasting \cite{ravuri2021skilful,pmlr-v235-gong24a}, often surpassing traditional numerical weather prediction models \cite{bauer2015quiet}. These methods are trained on fixed latitude-longitude gridded data, primarily using reanalysis datasets like ERA5 \cite{hersbach2020era5} as \emph{ground truth}, which integrate model outputs with multi-source observations through data assimilation \cite{carrassi2018data}. The success of these deep learning approaches can be attributed to their powerful non-linear fitting capabilities, enabling them to minimize the deviation between predictions and the supervised gridded analysis fields.

Despite achieving such impressive prediction results, recent studies have revealed that grid-based data-driven weather forecasting models exhibit systematic biases when interpolated and verified against in situ observations on the off-grid station scale \cite{ramavajjala2023verification,yang2024multi,han2024fengwu}. 
Such systematic biases exist consistently across various gridded data products, including reanalysis products and numerical model forecast outputs \cite{wu2023interpretable}, which manifest themselves as forecast inaccuracies in practical human activities. 


Previous approaches to mitigate grid-to-station biases primarily focus on enhancing the spatial resolution of meteorological fields through downscaling \cite{sun2024deep}, which utilizes finer-resolution grids to represent the meteorological conditions of a given region. Although recent studies have demonstrated that higher-resolution meteorological fields can represent actual observations more accurately \cite{han2024fengwu}, systematic errors between regional-scale gridded analysis fields (which serve as downscaling ground truth) and in-situ observations remain unavoidable. This limitation stems from the fundamental mismatch between discrete grid-based modeling approaches and the inherently continuous spatial distribution of meteorological variables. Moreover, the training of downscaling models typically relies on high-resolution analysis fields for supervision. However, kilometer-scale (or even higher resolution) analysis fields are difficult to obtain publicly, and these supervision fields themselves still show significant deviations from actual observations \cite{zhao2024omg}.


\begin{figure}
    \centering
    \includegraphics[width=0.99\linewidth]{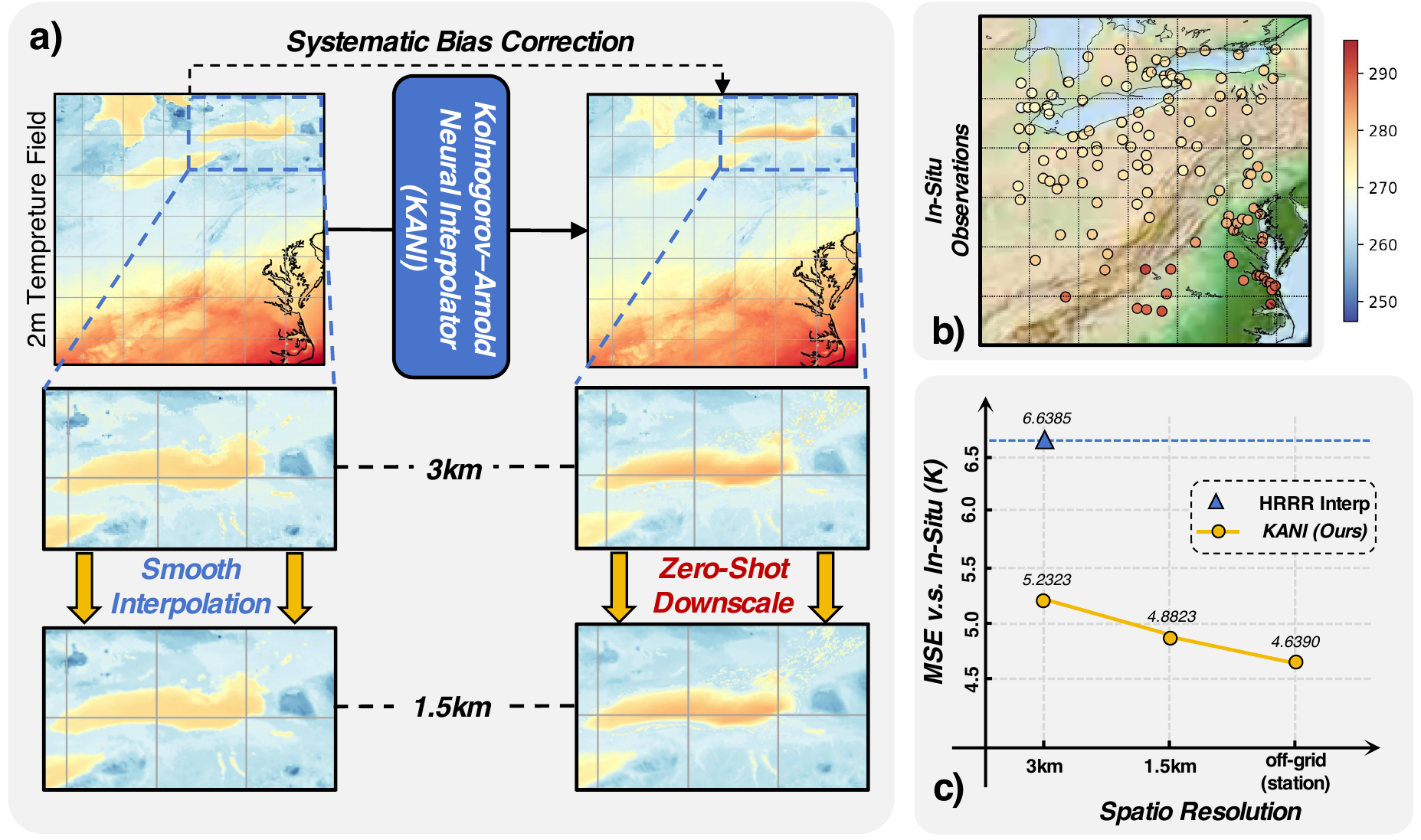}
    \caption{Visualization of systematic bias correction and zero-shot downscaling results using the proposed KANI method for input meteorological fields a). Through supervision from sparse in-situ observations b), KANI can reconstruct meteorological fields that are more consistent with actual observations, achieving zero-shot downscaling without requiring high-resolution grid supervision. Moreover, the validation accuracy against in-situ stations progressively improves with increasing resolution of the reconstructed fields, significantly exceeding the validation accuracy of the input fields c).}
    \label{fig:Intro}
    \vspace{-10pt} 
\end{figure}

In this paper, we propose a paradigm shift from traditional discrete grid-based representations to continuous neural field modeling of meteorological variables, which naturally aligns with their spatially continuous characteristics. By leveraging exclusively in-situ observations - the only truly accurate ground-truth measurements of atmospheric conditions - our approach constructs accurate and continuous meteorological fields that can be queried at arbitrary spatial locations, eliminating the reliance on high-resolution analysis fields for supervision. This continuous representation enables \emph{zero-shot} arbitrary resolution downscaling and systematic bias correction for input meteorological fields (as shown in Figure \ref{fig:Intro}), fundamentally addressing the limitations of discrete grid-based methods.

To achieve this, we propose the \emph{Kolmogorov–Arnold Neural Interpolator (KANI)}, a theoretically-grounded architecture inspired by the Kolmogorov–Arnold representation theorem, which guarantees the existence of continuous universal approximators for multivariate functions. Built upon the hypernetwork framework \cite{chauhan2024brief}, KANI comprises three key components: a meteorological field convolutional \emph{encoder}, a MLP-based weight \emph{generator}, and a KAN-based \cite{liu2024kan} neural radiance field \emph{reconstructor}. The encoder first extracts features from input gridded meteorological fields to guide the generator in producing partial weight parameters. Leveraging the Kolmogorov–Arnold theorem, our KAN-based reconstructor learns a continuous mapping from multi-scale coordinates to meteorological states, seamlessly bridging scales from subgrid-station to various grid resolutions while incorporating topographical information for enhanced spatial generalization. Once trained, KANI can recover and correct accurate meteorological states by learning the distribution of historical systematic biases, effectively generalizing across diverse geographical locations and spatial resolutions in accordance with topographical features.

To validate the effectiveness of the proposed method, we conducted extensive experiments across three study regions in the continental United States (CONUS), using HRRR 3km analysis data \cite{dowell2022high} as input meteorological fields and incorporating multiscale terrain information. The model was supervised using Weather-5K \cite{han2024weather} in-situ observations as ground truth, with multiple baseline methods designed for comparison. Our experimental results demonstrate KANI's superior performance, achieving remarkable accuracy improvements of 40.28\% for temperature and 67.41\% for wind speed predictions. Through comprehensive quantitative and qualitative analyses, we further verify that the model maintains high accuracy at station scale while exhibiting strong generalization capabilities across multiple grid resolutions.
    


\vspace{-5pt}
\section{Related Work}
\subsection{Weather Forecast Post-Processing with Deep Learning}
Post-processing of weather forecasts constitutes a crucial component in operational meteorological forecasting systems, aiming to enhance the spatiotemporal resolution of forecast results and mitigate inherent systematic errors, commonly called downscaling and bias correction, respectively.

Specifically, downscaling, analogous to super-resolution (SR) techniques in computer vision, involves the transformation of coarse-resolution global-scale meteorological fields into fine-resolution regional-scale gridded representations \cite{sun2024deep}. 
Various deep learning architectures, including Convolutional Neural Networks (CNNs) \cite{vandal2017deepsd}, Transformers \cite{zhong2024investigating}, Generative Adversarial Networks (GANs) \cite{leinonen2020stochastic}, Diffusions \cite{srivastava2024precipitation}, and Mamba \cite{liu2024mambads} models, have been extensively employed for downscaling applications.
Within the super-resolution modeling framework, while supervision using high-resolution meteorological fields is a prerequisite, such data (particularly meteorological analysis fields with resolution finer than 3km) is generally not readily accessible through public channels. Furthermore, the downscaled outputs are typically confined to predefined fixed grid resolutions, exhibiting limited capacity for effective cross-scale generalization \cite{liu2024deriving}.
To overcome these limitations, we propose a novel approach that leverages sparse in-situ observational supervision in conjunction with implicit neural representations of meteorological fields and accessible high-resolution terrain information. This approach facilitates zero-shot downscaling at arbitrary resolutions, enabling adaptive retrieval of accurate meteorological states across scales, from arbitrary grid scales down to off-grid scales, in accordance with practical demands.

With respect to bias correction, existing studies have predominantly addressed systematic biases inherent in forecast models \cite{lafferty2023downscaling,wu2024weathergnn}, developing bias correction models to capture and learn the systematic discrepancies between forecast outputs and reanalysis data. Unlike previous work, this paper primarily focuses on the systematic errors between gridded meteorological fields and actual in-situ observations, which represent persistent and inevitable biases inherent to traditional fixed-grid representations of meteorological variables. 
\vspace{-5pt}
\subsection{Kolmogorov–Arnold Networks}
As an alternative to multilayer perceptrons (MLPs), Kolmogorov–Arnold Networks (KAN) \cite{liu2024kan}, a novel neural network inspired by the Kolmogorov–Arnold representation theorem, has recently received significant attention.
In contrast to MLPs which learn weight parameters on edges while using fixed activation functions on nodes, KAN implements nonlinear modeling of complex systems through learnable spline-parametrized univariate functions, enabling a more flexible and interpretable framework for function approximation \cite{somvanshi2024survey}. Thanks to its unique architecture, KAN has been widely applied across various fields including computer vision \cite{li2024u} and AI4Science \cite{liu2024kan1}.

In this paper, we aim to leverage KAN's powerful function approximation capability to enhance the continuous representation of meteorological fields. 
Specifically, we integrate KAN into a hypernetwork-based implicit neural representation framework and construct the target network architecture using a hybrid KAN-MLP structure. Experimental results show that incorporating the KAN architecture not only improves the model's fitting capability while reducing the number of parameters but also accelerates the model's convergence rate.

\vspace{-5pt}
\section{Method}
\subsection{Problem Formulation}
Given a gridded meteorological field ${\boldsymbol x} \in \mathbb{R}^{1\times w\times h}$ with grid resolution $r$ (e.g. $r=3km$ for HRRR analysis fields \cite{dowell2022high}), where the latitude-longitude coordinates of each pixel are denoted as $\text{Coord}_{\text{grid\_r}} \in \mathbb{R}^{2\times w\times h}$.
We aim to train a neural network model to achieve continuous implicit neural representation of input gridded meteorological field while correcting the systematic biases between the field and in-situ observations.

Specifically, the corrected continuous representation of the input field ${\boldsymbol x}$ can be described as $f_{\theta}({\boldsymbol x})$. Based on this foundation, we can index the corresponding meteorological states by inputting different forms of coordinates. For example, by inputting grid coordinates at the original resolution r, we can achieve systematic bias correction at the original grid resolution:
\begin{equation}
    \hat{\boldsymbol y}_\text{correct} = f_{\theta}({\boldsymbol x}|\text{Coord}_{\text{grid\_r}}).
\end{equation}
Furthermore,  downscaling can be achieved through the input of grid coordinates at finer resolutions:
\begin{equation}
    \hat{\boldsymbol y}_\text{downscale} = f_{\theta}({\boldsymbol x}|\text{Coord}_{\text{grid\_r'}}).
\end{equation}
where $\text{Coord}_{\text{grid\_r'}}\in\mathbb{R}^{2\times w'\times h'}$, and $w'>w, h'>h, r'>r$. Additionally, we can obtain meteorological states at arbitrary locations by inputting target $n$ station coordinates $\text{Coord}_{\text{off\_grid}}\in\mathbb{R}^{2\times n}$, achieving what is known as station-scale downscaling \cite{liu2024deriving}:
\begin{equation}
    \hat{\boldsymbol y}_\text{off\_grid} = f_{\theta}({\boldsymbol x}|\text{Coord}_{\text{off\_grid}}).
\end{equation}
For an ideal continuous representation, the station interpolation error of model outputs should decrease as resolution increases, reaching optimality under direct sampling conditions (as shown in Figure \ref{fig:Intro}c)). Simultaneously, increased grid resolution should enable the representation of richer textural details.

\vspace{-5pt}
\subsection{Kolmogorov–Arnold Neural Interpolator}
\begin{figure*}
    \centering
    \includegraphics[width=0.9\linewidth]{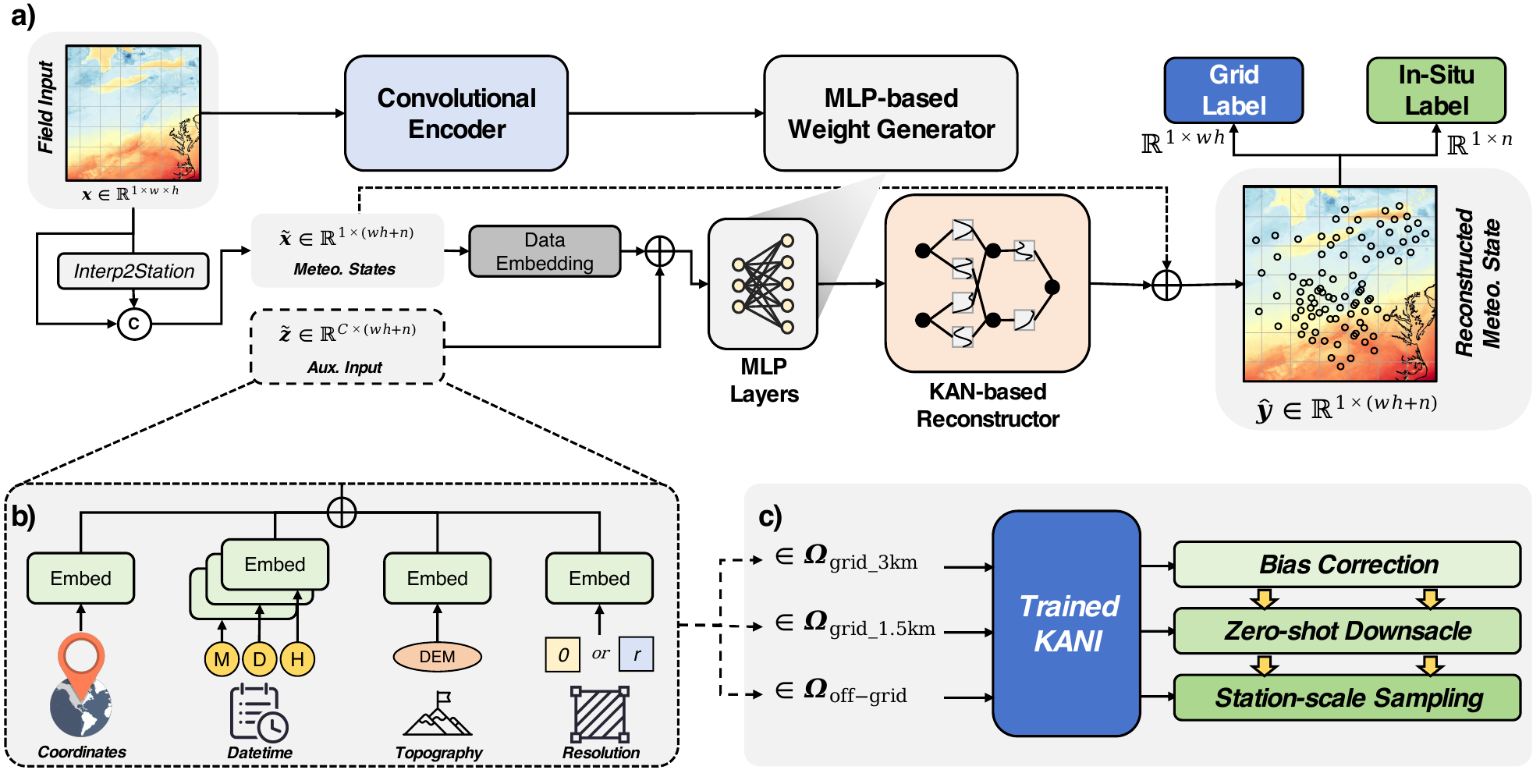}
    \caption{Illustration of our proposed Kolmogorov–Arnold Neural Interpolator (KANI) architecture a), which mainly consists of three parts: a convolutional encoder to encode the input grid meteorological field. A weight generator produces the weights of the target MLP layers based on the high-level features extracted from the input field. And the KAN-based reconstructor to learn the nonlinear mapping between the embedded input both from meteorological states and auxiliary inputs b) and corresponding reconstruct states at the grid scale and the station scale. Once trained, KANI can accomplish different target tasks by inputting auxiliary information at different scales c). }
    \label{fig:model}
    \vspace{-10pt} 
\end{figure*}
\noindent\textbf{Overall Structure:}
We designed a new model called Kolmogorov–Arnold Neural Interpolator (KANI) to achieve continuous representation and systematic bias correction of input gridded meteorological fields. As shown in Figure \ref{fig:model} a), the overall architecture of KANI is a hypernetwork model \cite{chauhan2024brief}, which consists of three components: a convolutional encoder, a weight generator, and a reconstruction.
Two types of input data are fed into KANI: the gridded meteorological fields are input into the convolutional encoder to extract high-level semantic features, which are then fed into the weight generator to produce partial weights for the reconstruction. 
The reconstructor takes multiple types of auxiliary information as input, including coordinates, dates, topography, and resolution (as shown in Figure \ref{fig:model} b)), as well as the interpolated meteorological states at both grid and station scales.
The optimization of model parameters is achieved through both self-supervision from original grid data and supervision from in-situ observations of the output meteorological states at grid and station scales, with training conducted in an end-to-end manner.

\noindent\textbf{Grid Meteorological Field Feature Extraction:}
For the input field ${\boldsymbol x}\in\mathbb{R}^{1\times w\times h}$, we first use a UNet-like architecture based on ResNet-18 \cite{he2016deep} to generate high-level semantic features. Specifically, we utilize the output from the 4th block of the UNet decoder as the extracted features with the dimension of ${\boldsymbol a}\in\mathbb{R}^{c\times\frac{w}{16}\times\frac{h}{16}}$. Then, we flatten the features into ${\boldsymbol a}\in\mathbb{R}^{ c\times d}, d\triangleq \frac{w}{16}\times\frac{h}{16}$ and fed into the weight generator. The generator contains two sub-modules used to generate the weight parameters for two MLP layers. Each weight can be computed by:
\begin{equation}
    w_i = \mathrm{Linear}(\mathrm{ReLU}(\mathrm{Conv1d({\boldsymbol a})})), i=1,2
\end{equation}
The above computational process transforms the input grid meteorological field into weight variables containing semantic information, thereby providing prior information for the reconstructor's learning.

\noindent\textbf{KAN-based Reconstructor:}
Inspired by implicit neural representations and neural radiance field methods in the continuous representation of 3D shapes and other objects \cite{xie2022neural}, we developed a specialized design that incorporates the characteristics of meteorological variable fields and sparse in-situ observations to reconstruct continuous-resolution representations from gridded meteorological fields. 
Specifically, we made improvements and innovations in two aspects: input and model structure.

The reconstructor's input primarily consists of two types of data inputs and one weight input. First, based on the original gridded meteorological field, we obtained the meteorological states of in-situ observations through interpolation and concatenated them with gridded meteorological observations to serve as the meteorological state input $\tilde{\boldsymbol {x}}\in \mathbb{R}^{1\times(wh+n)}$. Additionally, as shown in Figure \ref{fig:model}b), we incorporated other auxiliary inputs at both grid and station scales to guide the construction of continuous representation. This includes latitude and longitude coordinates, time, topography information, and resolution information Among these, topography information includes the gridded DEM field and the actual elevation values of each in-situ observation station. Regarding resolution information, the grid points were assigned their true resolution value $r$, while in-situ stations were assigned a resolution of $0$, with these values being broadcast across corresponding dimensions. All inputs mentioned above are encoded into embedding dimension $C$ through independent fully connected layers and then summed to form the final auxiliary input $\tilde{\boldsymbol {z}}\in\mathbb{R}^{C\times(wh+n)}$.

Subsequently, the constructed input first passes through two fully connected layers, whose weights are generated by the weight generator:
\begin{align}
        \tilde{\boldsymbol {b}} &= w_1 \cdot (\tilde{\boldsymbol {z}} + \mathrm{Embed}(\tilde{\boldsymbol x}))^\top\\
    \tilde{\boldsymbol {b}} &= \mathrm{ReLU}(\mathrm{LayerNorm}(\tilde{\boldsymbol {b}}))\\
    \tilde{\boldsymbol {b}} &= w_2 \cdot\tilde{\boldsymbol {h}}\\
    \tilde{\boldsymbol {b}} &= \mathrm{ReLU}(\mathrm{LayerNorm}(\tilde{\boldsymbol {b}}))
\end{align}
where the output hidden state $\tilde{\boldsymbol {b}}\in\mathbb{R}^{(wh+n)\times C'}$ and $\mathrm{Embed(\cdot)}$ represents a learnable linear embedding.

Unlike previous works that used MLP as the backbone network for implicit neural representation \cite{mildenhall2021nerf}, we employ the KAN structure \cite{liu2024kan} as the core architecture of the reconstruction. In contrast to MLP which optimizes node weights using fixed activation functions, KAN trains multiple B-spline parameters to approximate complex activation functions and aggregates activation values through direct summation, substantially reducing the model's node count while preserving strong nonlinear approximation capabilities. Therefore, we aim to utilize KAN's characteristics to improve both the computational efficiency and expressive power of the reconstruction. 
Consequently, our architecture first employs a fully connected layer for feature dimensionality reduction, followed by a KAN layer for further feature encoding, and ultimately utilizes an MLP layer to generate the final reconstruction output:
\begin{align}
\tilde{\boldsymbol {b}} &= \mathrm{Linear}(\tilde{\boldsymbol {b}})\\
\tilde{\boldsymbol {b}} &= (\Phi_{L-1}\circ\cdots\circ\Phi_1\circ\Phi_0)\tilde{\boldsymbol {b}}\\
\hat{\boldsymbol {y}} &= \mathrm{Linear}(\tilde{\boldsymbol {b}})^\top + \tilde{\boldsymbol {x}}
\end{align}
where $\Phi_i=\{\phi_{p,q}\}_i, p=1,\ldots, C_{in}, q=1,\ldots, C_{out}$ and $\phi(\cdot)$ is a function which has trainable parameters described by spline function. The output $\hat{\boldsymbol {y}}\in\mathbb{R}^{1\times(wh+n)}$ represents the reconstruction states at both grid and station points.

\noindent\textbf{Loss Function:}
For the KANI output, we use both the original resolution grid data and in-situ observations for supervision, with mean square error as the loss function:
\begin{equation}
    \begin{split}
        \mathcal{L} &= \mathcal{L}_{\mathrm{grid}}+\mathcal{L}_{\mathrm{station}}\\
        &=\|\boldsymbol{y}_{\mathrm{grid}}-\hat{\boldsymbol{y}}_{\mathrm{grid}}\|^2 + \|\boldsymbol{y}_{\mathrm{station}}-\hat{\boldsymbol{y}}_{\mathrm{station}}\|^2
    \end{split}
\end{equation}
Notably, although prior studies have addressed the conflict between station-based and grid-based losses \cite{liu2024deriving}, our approach mitigates this contradiction to a certain degree by incorporating distinct resolution embeddings for grid points and station points respectively. The joint constraint of both losses enables the model to reconstruct the original meteorological field while correcting systematic biases at sub-grid scales through station supervision, and generalizing to arbitrary spatial locations guided by topography texture.
    
\subsection{Zero-Shot Arbitrary-Scale Reconstruction}
As shown in Figure \ref{fig:model}c), once KANI is trained, during the inference phase, we set the input resolution to 0 to correct systematic biases in the original meteorological field at off-grid scales. Additionally, by inputting higher-resolution topography data and grid coordinates, we can achieve zero-shot downscaling of meteorological fields at arbitrary grid scales guided by high-resolution topography texture, without requiring supervision from high-resolution analysis fields.
The guidance from topography features has strong physical significance, particularly for near-surface meteorological variables.

\vspace{-5pt}
\section{Experiment}
\begin{figure}
    \centering
    \includegraphics[width=0.99\linewidth]{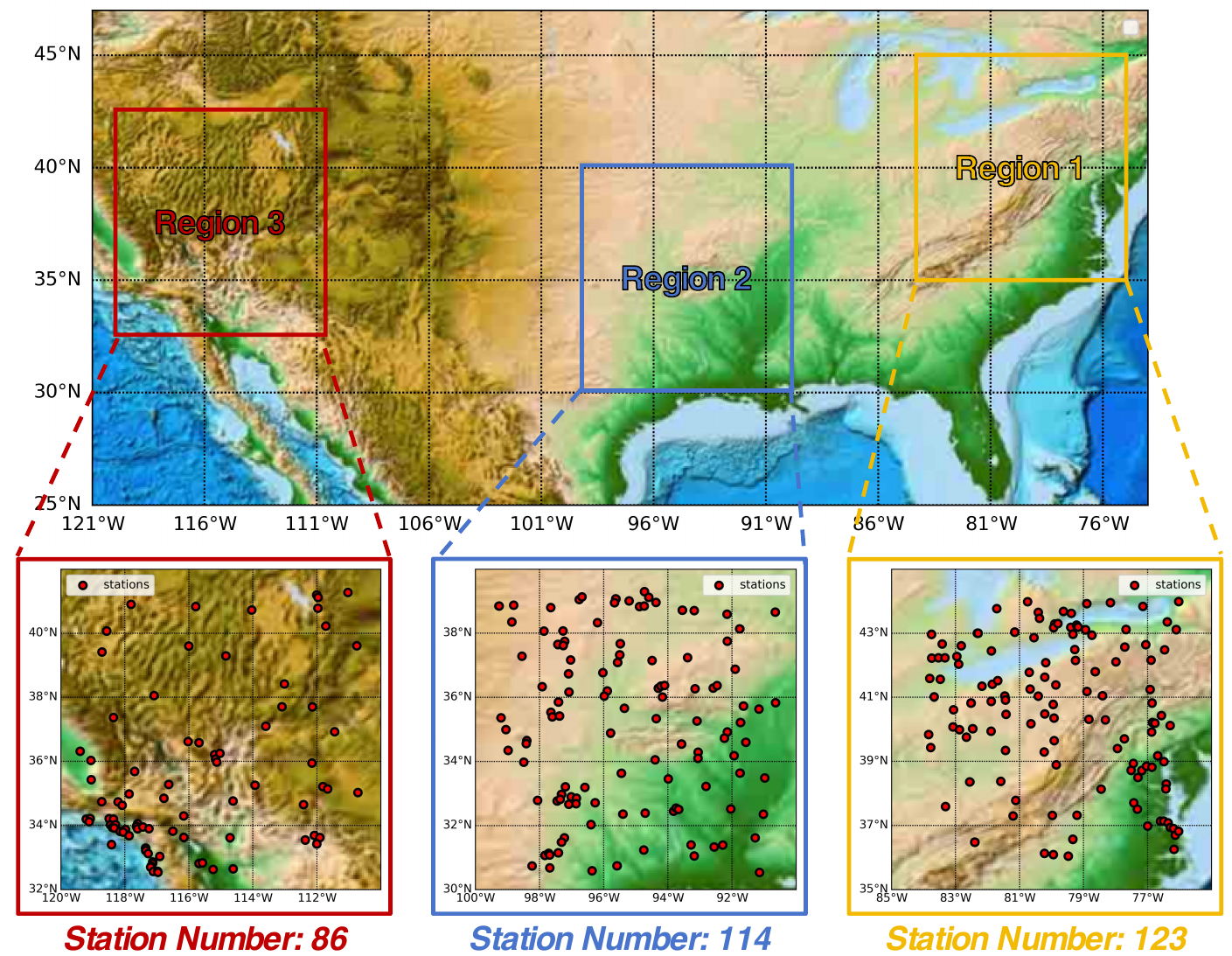}
    \caption{The study area in this paper and the distribution of corresponding in-situ observations.}
    \label{fig:study area}
    \vspace{-10pt} 
\end{figure}
\subsection{Experiment Setting}
\subsubsection{Study Area and Dataset Description}
\begin{figure*}
    \centering
    \includegraphics[width=0.99\linewidth]{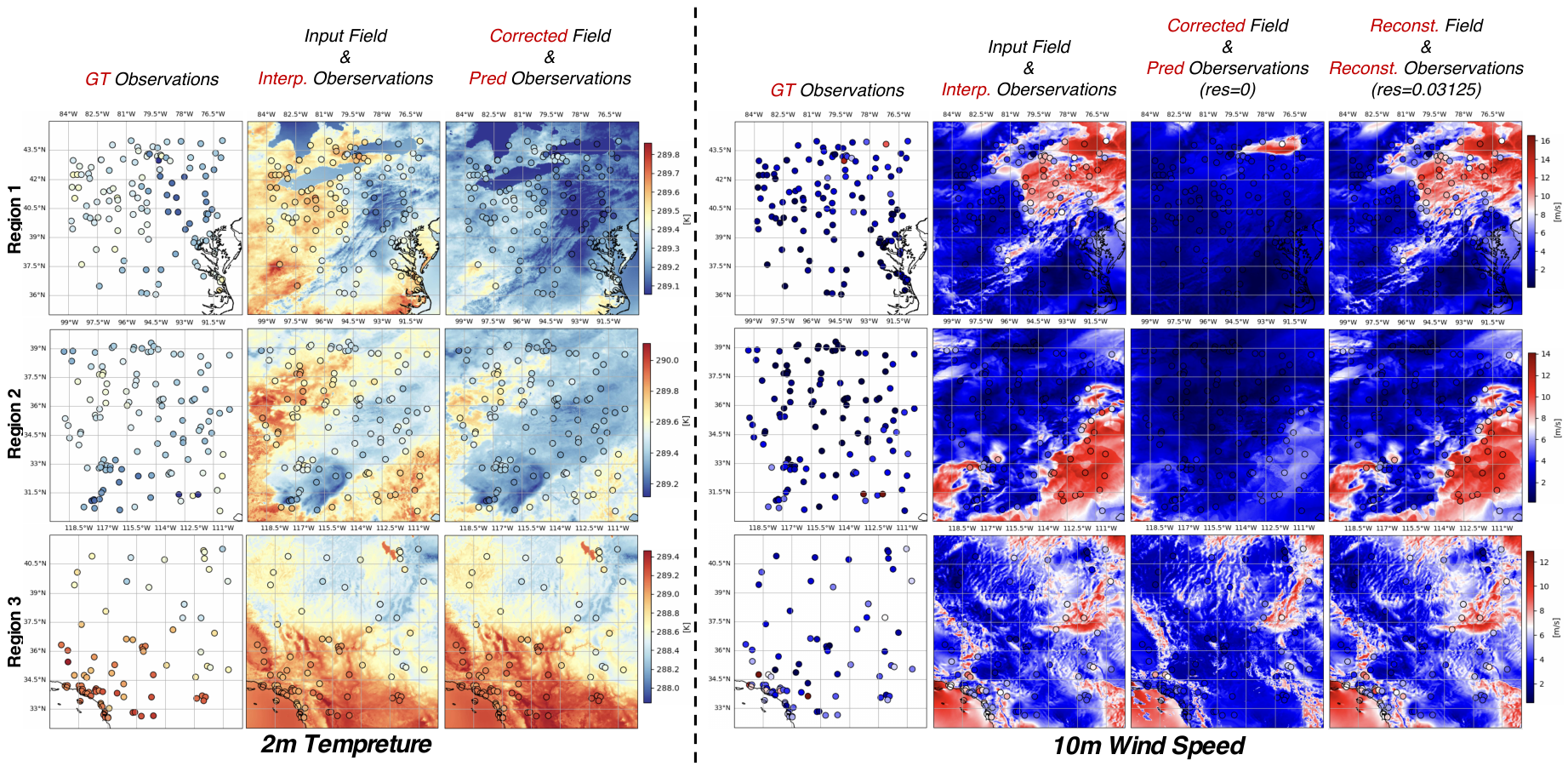}
    \caption{The illustration of proposed KANI for correcting the grid-station systematic bias for the input meteorological field. As shown in the figure, significant biases (different color distributions) exist between the input fields and their interpolated observations presented in the second column compared to the ground truth observations shown in the first column. Through KANI's correction, we achieve predictions more consistent with ground truth at in-situ station locations and generalize these corrections across the entire field based on topography texture patterns, thereby accomplishing the goal of bias correction.}
    \label{fig:BC}
\end{figure*}

To validate the effectiveness of the proposed KANI, as shown in Figure \ref{fig:study area}, we selected three sub-regions within the continental United States (CONUS) as study areas. Each region covers a $10^\circ\times10^\circ$ latitude-longitude range and includes various typical terrain features such as mountains and plains. Within each sub-region, we extracted the weather stations contained in the region from the Weather-5K dataset \cite{han2024weather}. For meteorological field data, we selected HRRR analysis \cite{dowell2022high} fields as the input gridded meteorological fields for the model and interpolated the original data onto a latitude-longitude grid with a 3km resolution (i.e. $320\times320$ pixels for each sub-region).
To align station and gridded data, we selected two of the most common surface meteorological variables as our research subjects: 2-meter temperature ($t_{2m}$) and 10-meter wind speed ($gust$). It should be noted that there is a difference between the observed values in Weather-5K and the analyzed values in HRRR for wind speed variables: the former measures instantaneous wind speed, while the latter represents the maximum gust speed within an hour. Therefore, for wind speed variables, the mapping learned by the model includes not only the grid-station systematic but also the reconstruction from maximum gust speed to instantaneous wind speed.
Obtaining high-resolution states of these two variables has significant practical implications for downstream meteorological tasks \cite{zhong2024investigating}. We selected hourly data from 2017 to 2021 (5 years) and divided the data into training, validation, and test sets. For more information please refer to Appendix \ref{sec:Dataset}.

\vspace{-5pt}
\subsubsection{Training Details}
\vspace{-5pt}

We follow \cite{liu2024mambads} to train the model for 120 epochs and optimize the model using the Adam with $\beta_1=0.9, \beta_2=0.999$, and employ a step learning rate initialized as 1e-4 and decreased by half once the number of epochs reaches one of the milestones $(60, 96,108, 114)$. We trained our proposed model using 4x NVIDIA A100-40G GPUs, setting the batch size to 32 per GPU.

\subsection{Bias Correction from In-Situ Observations}
Correcting the systematic bias between gridded meteorological fields and in-situ field observations is the primary task of this paper. Therefore, through visualization, we first demonstrate the bias correction of the input meteorological field by the proposed KANI. Figure \ref{fig:BC} shows the bias correction effects of KANI on temperature and wind speed fields across different regions. 

Specifically, for 2-m temperature, the figure shows significant discrepancies between in-situ observations (GT observations) and the input temperature field. For example, in Region 1, the observed values in the central and western areas are notably lower than the assimilated temperature field results. Through the correction by the proposed KANI model, the predicted station observations are significantly closer to the in-situ observations. More importantly, KANI can \emph{generalize the correction results from observation stations to other grid locations based on topography texture}, even where no observation stations are deployed. This indicates that KANI can reconstruct meteorological fields that better match real observations at the original scale through the constraints of sparse observation stations, which has a significant positive impact on improving the actual forecast accuracy of weather prediction models.

For the 10-m wind speed variable, we present prediction results from two different input settings. This is attributed to different observation windows between in-situ observations and meteorological analysis field data, where the former focuses on instantaneous values while the latter concerns maximum gust values within an hour. Consequently, the observed measurements are notably lower than the results from meteorological fields. Under this context, the model learns the correction process from hourly windows to instantaneous values. As demonstrated in the third column of results, KANI effectively reconstructs instantaneous wind fields from gust fields according to the distribution of instantaneous observations, while preserving the original texture information in the reconstructed fields. Moreover, as shown by the results in the fourth column, by inputting resolution values at the original scale into the trained KANI, we can still recover meteorological field reconstructions consistent with gust distributions without additional training. This indicates that once the model completes training, it incorporates meteorological field distributions at multiple scales, a capability achieved by introducing resolution as auxiliary information to guide model training.

\subsection{Topography Guided Zero-Shot Downscaling}

\begin{figure}
    \centering
    \includegraphics[width=0.95\linewidth]{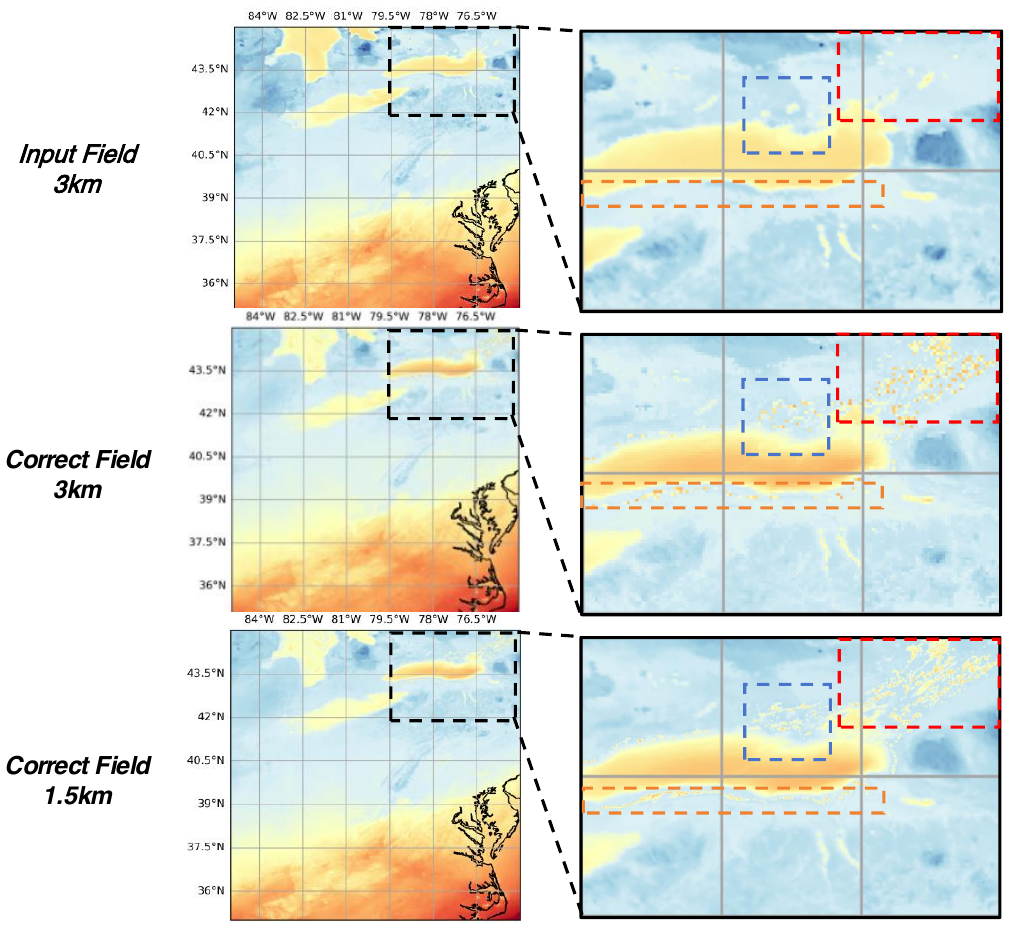}
    \caption{Illustration of the zero-shot downscaling result by inputting high-resolution coordinates and corresponding topography information into trained KANI. }
    \label{fig:DS}
    \vspace{-10pt} 
\end{figure}

In addition to reconstruction and bias correction at the original resolution, the proposed KANI exhibits a remarkable zero-shot downscaling capability. This is achieved by learning the continuous, nonlinear relationships between coordinate positions, topography, and meteorological states. As a result, the trained model can recover high-resolution spatial details by simply inputting high-resolution coordinate grids and corresponding topographic features, all without requiring supervision from high-resolution meteorological field data. As illustrated in Figure \ref{fig:DS}, starting from the input 3km 2m-temperature field, we obtain the corrected 3km field by feeding the original resolution geographical coordinates (latitude/longitude) and topography texture information into the trained KANI model. In this case, we can observe that more detailed textures have been restored. We attribute this to the low resolution of assimilated terrain information in the original data, and the fact that many sub-grid observational details were masked during the assimilation process.

Furthermore, by inputting higher-resolution coordinates and terrain texture data, we obtained downscaled results at 1.5km resolution. Compared to the 3km corrected results, the 1.5km downscaled outputs reveal enhanced textural details rather than merely smoothing coarse-resolution data. This outcome is encouraging, particularly since no high-resolution meteorological field data was used for supervision during the training process. This indicates that KANI has successfully learned the correlation between terrain texture and temperature, which is both physically explainable and empirically verified in real-world conditions. This quantitative visualization further validates KANI's exceptional performance in kilometer-scale downscaling using only sparse observational supervision, demonstrating both the feasibility and effectiveness of this new paradigm in transcending the traditional fixed-grid downscaling approach that relies on high-resolution supervision.

\subsection{Comparison with Baselines Against In-Situ Observations}

\begin{table*}[]
    \centering
    \renewcommand{\arraystretch}{1.1}
    \small
    \scalebox{0.9}{ 
    \begin{tabular}{c|cc|cc|cc|cc|cc|cc}
    \toprule
         {Region}&\multicolumn{4}{c|}{Region 1}&\multicolumn{4}{c|}{Region 2}&\multicolumn{4}{c}{Region 3}  \\\hline
         \multirow{2}{*}{\diagbox[width=10em,height=2.4em]{Model}{Variable}}&\multicolumn{2}{c|}{$t_{2m}$}&\multicolumn{2}{c|}{$gust$}&\multicolumn{2}{c|}{$t_{2m}$}&\multicolumn{2}{c|}{$gust$}&\multicolumn{2}{c|}{$t_{2m}$}&\multicolumn{2}{c}{$gust$}\\ 
         &MSE&MAE&MSE&MAE&MSE&MAE&MSE&MAE&MSE&MAE&MSE&MAE\\\hline
         Linear Interp.&6.6385&1.7370&15.3531&3.1024&5.3590&1.5480&13.9193&3.0746&23.0984&4.8765&8.3901&2.7847\\
         Nearest Interp.&6.9564&1.7709&15.3549&3.0988&5.3409&3.0947&13.9243&3.0778&23.0732&4.8688&8.3948&2.7904\\
         MLP&5.8749&1.7097&4.0387&2.5876&4.2567&1.4678&3.1423&1.4167&12.2914&3.0086&6.9476&1.7842\\
         HyperMLP&5.5883&1.6550&2.9926&1.2180&3.8567&1.3659&2.7937&1.2948&10.2034&2.9476&5.3745&1.5084\\
        KANI (Ours)&\textbf{4.6390}&\textbf{1.4616}&\textbf{2.8580}&\textbf{1.1776}&\textbf{3.6425}&\textbf{1.3283}&\textbf{2.6212}&\textbf{1.1432}&\textbf{9.5449}&\textbf{2.7845}&\textbf{5.0608}&\textbf{1.4487}\\
    \bottomrule
    \end{tabular}}
    \caption{The overall performance comparison for 2m-temperature ($t_{2m}$) and wind-speed ($gust$) of the proposed KANI and other baseline models against in-situ observations.}
    \label{tab:my_label}
\end{table*}

Mitigating the systematic bias between gridded meteorological fields and observation stations is the primary objective of this paper. Therefore, we compared the performance of our proposed method with other baseline models in terms of systematic bias at in-situ observation stations within different study areas.

\noindent\textbf{Evaluation Metrics:}
To verify the accuracy of the reconstructed meteorological states, we evaluated the errors between the outputs of different methods and the actual in-situ observation stations. Therefore, following previous studies on station forecasting \cite{han2024weather}, we used Mean Absolute Error (MAE) and Mean Square Error (MSE) as metrics to evaluate the overall performance of the model. 

\noindent\textbf{Baseline Setup:} Due to the current lack of existing methods in this research area, we established several basic baseline models for comparison, including direct interpolation of meteorological fields (including linear interpolation and nearest neighbor interpolation) and neural network approaches. The latter includes pure MLP and MLP-based hypernetwork structures. To ensure a fair comparison, we maintained consistent inputs, outputs, and optimization methods across all baseline approaches. Detailed model structures and designs of the baseline models are provided in the Appendix \ref{sec:Baseline}.

\noindent\textbf{Overall Performance:} As shown in Table \ref{tab:my_label}, direct interpolation of the input fields reveals substantial systematic bias between gridded field data and in-situ observations, with significant variations across different regions and variables. While all constructed baseline methods effectively correct the biases, it is evident that the hypernetwork-based structures of HyperMLP and KANI significantly outperform pure MLP architectures. This demonstrates the hypernetwork architecture's powerful capability in modeling complex nonlinear processes, resulting in enhanced modeling capacity and greater model expressiveness. Further comparison between HyperMLP and KANI reveals that the KAN-integrated KANI model significantly outperforms its purely MLP-structured counterpart, demonstrating KAN's robust capability in modeling highly nonlinear dynamical systems such as meteorological processes. As demonstrated in the Appendix \ref{sec:Baseline} Table \ref{tab:apd_comp}, a comparison of parameter counts and computational speeds across different methods further validates that the KANI model maintains high efficiency while achieving superior performance, featuring fewer parameters and inference speeds comparable to HyperMLP.
\vspace{-5pt}
\subsection{Generalization Capabilities Across Various Resolutions}

\begin{figure}
    \centering
    \includegraphics[width=0.99\linewidth]{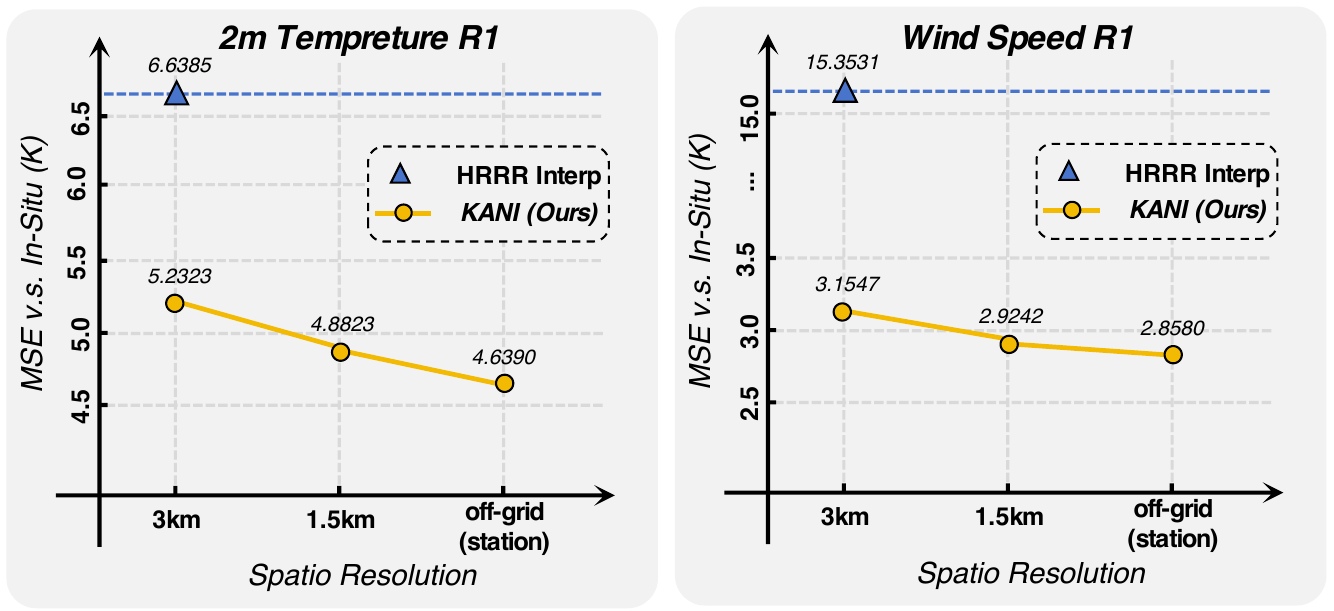}
    \caption{Accuracy comparison between interpolation results of KANI-reconstructed fields at different resolutions and the original meteorological field interpolation to in-situ stations.}
    \label{fig:scale}
    \vspace{-15pt} 
\end{figure}

To validate the rationality of KANI's reconstructed meteorological fields at different resolutions, we evaluated the interpolation accuracy of these fields when interpolated to in-situ observation stations. While previous results were based on direct sampling at observation points, this experiment assesses the consistency and usability of KANI's fields across resolutions. Figure \ref{fig:scale} shows the original HRRR fields exhibit consistently high interpolation errors, reflecting systematic biases between gridded data and observations. In contrast, KANI-reconstructed fields show progressively lower errors as resolution increases, with optimal performance achieved through direct sampling at stations.
The results also demonstrate KANI's strong generalization across resolutions, as the model maintains low interpolation errors at varying scales. This resolution-agnostic performance underscores KANI's ability to transcend traditional grid-based limitations, providing a robust framework for accurate meteorological field reconstruction and bias correction.
\vspace{-5pt}
\subsection{Evaluation Under Different Auxiliary Inputs}

\begin{table}[]
    \centering
    \renewcommand{\arraystretch}{1.1}
    \small
    \scalebox{0.9}{ 
    \begin{tabular}{c|cc|cc}
    \toprule
         {Region}&\multicolumn{4}{c}{Region 1}  \\\hline
         \multirow{2}{*}{\diagbox[width=10em,height=2.3em]{Model}{Variable}}&\multicolumn{2}{c|}{$t_{2m}$}&\multicolumn{2}{c}{$gust$}\\ 
         &MSE&MAE&MSE&MAE\\\hline
         KANI&\textbf{4.6390}&\textbf{1.4616}&\textbf{2.8580}&\textbf{1.1776}\\
         w/o Date&4.6422&1.4637&2.8605&1.1842\\
         w/o Topo.&4.8245&1.6048&2.9042&1.2086\\
         w/o Res.&4.9253&1.6843&2.9535&1.2123\\
    
    \bottomrule
    \end{tabular}}
    \caption{The overall performance comparison for 2m-temperature ($t_{2m}$) and wind-speed ($gust$) of the proposed KANI and other baseline models against in-situ observations.}
    \label{tab:ab_study}
    \vspace{-10pt} 
\end{table}

To evaluate the importance of auxiliary inputs, we conducted an ablation study comparing KANI's performance with and without key components such as topography, temporal, and resolution information. As shown in Table 2, the exclusion of topography information leads to a noticeable increase in MAE, from 1.46 to 1.68 for temperature, underscoring its critical role in capturing local meteorological variations. Similarly, removing resolution information results in an MAE rise from 1.46 to 1.63 for temperature, demonstrating its importance for scale-adaptive reconstruction. When all auxiliary inputs are included, KANI achieves the best performance, confirming the synergistic effect of these features in enhancing the model's accuracy.



\vspace{-5pt}
\section{Conclusion}
In this paper, we proposed the \emph{Kolmogorov-Arnold Neural Interpolator} (KANI), a novel framework that leverages the Kolmogorov-Arnold theorem to construct continuous neural representations of meteorological fields from discrete gridded data. By integrating sparse in-situ observations and high-resolution topographic information, KANI achieves zero-shot downscaling and systematic bias correction without requiring high-resolution supervision. The continuous neural representation enables KANI to align more closely with real-world observations, effectively mitigating the persistent grid-to-station systematic errors that are unavoidable in traditional methods. Our experiments demonstrate significant improvements in temperature and wind speed predictions, showcasing KANI's ability to transcend traditional grid-based limitations. This work offers a scalable and accurate solution for meteorological post-processing, particularly in regions with limited high-resolution data.




\section*{Impact Statement}

The Kolmogorov-Arnold Neural Interpolator (KANI) has the potential to significantly enhance the accuracy and reliability of weather forecasts, particularly in regions where high-resolution meteorological data is scarce. By enabling zero-shot downscaling and systematic bias correction, KANI can improve decision-making in critical sectors such as agriculture, aviation, and disaster management. Furthermore, the integration of topographic information provides a physically interpretable framework, bridging the gap between data-driven models and traditional meteorological methods. This work not only advances the field of meteorological post-processing but also opens new possibilities for applying continuous neural representations in other scientific domains.


\nocite{langley00}

\bibliography{example_paper}

\begin{thebibliography}{35}
\providecommand{\natexlab}[1]{#1}
\providecommand{\url}[1]{\texttt{#1}}
\expandafter\ifx\csname urlstyle\endcsname\relax
  \providecommand{\doi}[1]{doi: #1}\else
  \providecommand{\doi}{doi: \begingroup \urlstyle{rm}\Url}\fi

\bibitem[Bauer et~al.(2015)Bauer, Thorpe, and Brunet]{bauer2015quiet}
Bauer, P., Thorpe, A., and Brunet, G.
\newblock The quiet revolution of numerical weather prediction.
\newblock \emph{Nature}, 525\penalty0 (7567):\penalty0 47--55, 2015.

\bibitem[Bi et~al.(2023)Bi, Xie, Zhang, Chen, Gu, and Tian]{bi2023accurate}
Bi, K., Xie, L., Zhang, H., Chen, X., Gu, X., and Tian, Q.
\newblock Accurate medium-range global weather forecasting with 3d neural networks.
\newblock \emph{Nature}, 619\penalty0 (7970):\penalty0 533--538, 2023.

\bibitem[Carrassi et~al.(2018)Carrassi, Bocquet, Bertino, and Evensen]{carrassi2018data}
Carrassi, A., Bocquet, M., Bertino, L., and Evensen, G.
\newblock Data assimilation in the geosciences: An overview of methods, issues, and perspectives.
\newblock \emph{Wiley Interdisciplinary Reviews: Climate Change}, 9\penalty0 (5):\penalty0 e535, 2018.

\bibitem[Chauhan et~al.(2024)Chauhan, Zhou, Lu, Molaei, and Clifton]{chauhan2024brief}
Chauhan, V.~K., Zhou, J., Lu, P., Molaei, S., and Clifton, D.~A.
\newblock A brief review of hypernetworks in deep learning.
\newblock \emph{Artificial Intelligence Review}, 57\penalty0 (9):\penalty0 250, 2024.

\bibitem[Chen et~al.(2023)Chen, Han, Gong, Bai, Ling, Luo, Chen, Ma, Zhang, Su, et~al.]{chen2023fengwu}
Chen, K., Han, T., Gong, J., Bai, L., Ling, F., Luo, J.-J., Chen, X., Ma, L., Zhang, T., Su, R., et~al.
\newblock Fengwu: Pushing the skillful global medium-range weather forecast beyond 10 days lead.
\newblock \emph{arXiv preprint arXiv:2304.02948}, 2023.

\bibitem[Dowell et~al.(2022)Dowell, Alexander, James, Weygandt, Benjamin, Manikin, Blake, Brown, Olson, Hu, et~al.]{dowell2022high}
Dowell, D.~C., Alexander, C.~R., James, E.~P., Weygandt, S.~S., Benjamin, S.~G., Manikin, G.~S., Blake, B.~T., Brown, J.~M., Olson, J.~B., Hu, M., et~al.
\newblock The high-resolution rapid refresh (hrrr): An hourly updating convection-allowing forecast model. part i: Motivation and system description.
\newblock \emph{Weather and Forecasting}, 37\penalty0 (8):\penalty0 1371--1395, 2022.

\bibitem[Gong et~al.(2024)Gong, Bai, Ye, Xu, Liu, Dai, Yang, and Ouyang]{pmlr-v235-gong24a}
Gong, J., Bai, L., Ye, P., Xu, W., Liu, N., Dai, J., Yang, X., and Ouyang, W.
\newblock {C}as{C}ast: Skillful high-resolution precipitation nowcasting via cascaded modelling.
\newblock In \emph{Proceedings of the 41st International Conference on Machine Learning}, volume 235 of \emph{Proceedings of Machine Learning Research}, pp.\  15809--15822. PMLR, 21--27 Jul 2024.

\bibitem[Han et~al.(2024{\natexlab{a}})Han, Guo, Chen, Xu, and Bai]{han2024weather}
Han, T., Guo, S., Chen, Z., Xu, W., and Bai, L.
\newblock Weather-5k: A large-scale global station weather dataset towards comprehensive time-series forecasting benchmark.
\newblock \emph{arXiv preprint arXiv:2406.14399}, 2024{\natexlab{a}}.

\bibitem[Han et~al.(2024{\natexlab{b}})Han, Guo, Ling, Chen, Gong, Luo, Gu, Dai, Ouyang, and Bai]{han2024fengwu}
Han, T., Guo, S., Ling, F., Chen, K., Gong, J., Luo, J., Gu, J., Dai, K., Ouyang, W., and Bai, L.
\newblock Fengwu-ghr: Learning the kilometer-scale medium-range global weather forecasting.
\newblock \emph{arXiv preprint arXiv:2402.00059}, 2024{\natexlab{b}}.

\bibitem[He et~al.(2016)He, Zhang, Ren, and Sun]{he2016deep}
He, K., Zhang, X., Ren, S., and Sun, J.
\newblock Deep residual learning for image recognition.
\newblock In \emph{Proceedings of the IEEE conference on computer vision and pattern recognition}, pp.\  770--778, 2016.

\bibitem[Hersbach et~al.(2020)Hersbach, Bell, Berrisford, Hirahara, Hor{\'a}nyi, Mu{\~n}oz-Sabater, Nicolas, Peubey, Radu, Schepers, et~al.]{hersbach2020era5}
Hersbach, H., Bell, B., Berrisford, P., Hirahara, S., Hor{\'a}nyi, A., Mu{\~n}oz-Sabater, J., Nicolas, J., Peubey, C., Radu, R., Schepers, D., et~al.
\newblock The era5 global reanalysis.
\newblock \emph{Quarterly Journal of the Royal Meteorological Society}, 146\penalty0 (730):\penalty0 1999--2049, 2020.

\bibitem[Huang et~al.(2024)Huang, Gianinazzi, Yu, Dueben, and Hoefler]{pmlr-v235-huang24h}
Huang, L., Gianinazzi, L., Yu, Y., Dueben, P.~D., and Hoefler, T.
\newblock {D}iff{DA}: a diffusion model for weather-scale data assimilation.
\newblock In \emph{Proceedings of the 41st International Conference on Machine Learning}, volume 235 of \emph{Proceedings of Machine Learning Research}, pp.\  19798--19815. PMLR, 21--27 Jul 2024.

\bibitem[Lafferty \& Sriver(2023)Lafferty and Sriver]{lafferty2023downscaling}
Lafferty, D.~C. and Sriver, R.~L.
\newblock Downscaling and bias-correction contribute considerable uncertainty to local climate projections in cmip6.
\newblock \emph{npj Climate and Atmospheric Science}, 6\penalty0 (1):\penalty0 158, 2023.

\bibitem[Leinonen et~al.(2020)Leinonen, Nerini, and Berne]{leinonen2020stochastic}
Leinonen, J., Nerini, D., and Berne, A.
\newblock Stochastic super-resolution for downscaling time-evolving atmospheric fields with a generative adversarial network.
\newblock \emph{IEEE Transactions on Geoscience and Remote Sensing}, 59\penalty0 (9):\penalty0 7211--7223, 2020.

\bibitem[Li et~al.(2024)Li, Liu, Li, Wang, Liu, Liu, Chen, and Yuan]{li2024u}
Li, C., Liu, X., Li, W., Wang, C., Liu, H., Liu, Y., Chen, Z., and Yuan, Y.
\newblock U-kan makes strong backbone for medical image segmentation and generation.
\newblock \emph{arXiv preprint arXiv:2406.02918}, 2024.

\bibitem[Liu et~al.(2024{\natexlab{a}})Liu, Chen, Bai, Li, Chen, Wang, Ouyang, Zou, and Shi]{liu2024deriving}
Liu, Z., Chen, H., Bai, L., Li, W., Chen, K., Wang, Z., Ouyang, W., Zou, Z., and Shi, Z.
\newblock Deriving accurate surface meteorological states at arbitrary locations via observation-guided continous neural field modeling.
\newblock \emph{IEEE Transactions on Geoscience and Remote Sensing}, 2024{\natexlab{a}}.

\bibitem[Liu et~al.(2024{\natexlab{b}})Liu, Chen, Bai, Li, Ouyang, Zou, and Shi]{liu2024mambads}
Liu, Z., Chen, H., Bai, L., Li, W., Ouyang, W., Zou, Z., and Shi, Z.
\newblock Mambads: Near-surface meteorological field downscaling with topography constrained selective state space modeling.
\newblock \emph{IEEE Transactions on Geoscience and Remote Sensing}, 2024{\natexlab{b}}.

\bibitem[Liu et~al.(2024{\natexlab{c}})Liu, Ma, Wang, Matusik, and Tegmark]{liu2024kan1}
Liu, Z., Ma, P., Wang, Y., Matusik, W., and Tegmark, M.
\newblock Kan 2.0: Kolmogorov-arnold networks meet science.
\newblock \emph{arXiv preprint arXiv:2408.10205}, 2024{\natexlab{c}}.

\bibitem[Liu et~al.(2024{\natexlab{d}})Liu, Wang, Vaidya, Ruehle, Halverson, Solja{\v{c}}i{\'c}, Hou, and Tegmark]{liu2024kan}
Liu, Z., Wang, Y., Vaidya, S., Ruehle, F., Halverson, J., Solja{\v{c}}i{\'c}, M., Hou, T.~Y., and Tegmark, M.
\newblock Kan: Kolmogorov-arnold networks.
\newblock \emph{arXiv preprint arXiv:2404.19756}, 2024{\natexlab{d}}.

\bibitem[Mayer et~al.(2018)Mayer, Jakobsson, Allen, Dorschel, Falconer, Ferrini, Lamarche, Snaith, and Weatherall]{mayer2018nippon}
Mayer, L., Jakobsson, M., Allen, G., Dorschel, B., Falconer, R., Ferrini, V., Lamarche, G., Snaith, H., and Weatherall, P.
\newblock The nippon foundation—gebco seabed 2030 project: The quest to see the world’s oceans completely mapped by 2030.
\newblock \emph{Geosciences}, 8\penalty0 (2):\penalty0 63, 2018.

\bibitem[Mildenhall et~al.(2021)Mildenhall, Srinivasan, Tancik, Barron, Ramamoorthi, and Ng]{mildenhall2021nerf}
Mildenhall, B., Srinivasan, P.~P., Tancik, M., Barron, J.~T., Ramamoorthi, R., and Ng, R.
\newblock Nerf: Representing scenes as neural radiance fields for view synthesis.
\newblock \emph{Communications of the ACM}, 65\penalty0 (1):\penalty0 99--106, 2021.

\bibitem[Ramavajjala \& Mitra(2023)Ramavajjala and Mitra]{ramavajjala2023verification}
Ramavajjala, V. and Mitra, P.~P.
\newblock Verification against in-situ observations for data-driven weather prediction.
\newblock \emph{arXiv preprint arXiv:2305.00048}, 2023.

\bibitem[Ravuri et~al.(2021)Ravuri, Lenc, Willson, Kangin, Lam, Mirowski, Fitzsimons, Athanassiadou, Kashem, Madge, et~al.]{ravuri2021skilful}
Ravuri, S., Lenc, K., Willson, M., Kangin, D., Lam, R., Mirowski, P., Fitzsimons, M., Athanassiadou, M., Kashem, S., Madge, S., et~al.
\newblock Skilful precipitation nowcasting using deep generative models of radar.
\newblock \emph{Nature}, 597\penalty0 (7878):\penalty0 672--677, 2021.

\bibitem[Somvanshi et~al.(2024)Somvanshi, Javed, Islam, Pandit, and Das]{somvanshi2024survey}
Somvanshi, S., Javed, S.~A., Islam, M.~M., Pandit, D., and Das, S.
\newblock A survey on kolmogorov-arnold network.
\newblock \emph{arXiv preprint arXiv:2411.06078}, 2024.

\bibitem[Srivastava et~al.(2024)Srivastava, Yang, Kerrigan, Dresdner, McGibbon, Bretherton, and Mandt]{srivastava2024precipitation}
Srivastava, P., Yang, R., Kerrigan, G., Dresdner, G., McGibbon, J.~J., Bretherton, C.~S., and Mandt, S.
\newblock Precipitation downscaling with spatiotemporal video diffusion.
\newblock In \emph{The Thirty-eighth Annual Conference on Neural Information Processing Systems}, 2024.

\bibitem[Sun et~al.(2024)Sun, Deng, Ren, Liu, Deng, and Jin]{sun2024deep}
Sun, Y., Deng, K., Ren, K., Liu, J., Deng, C., and Jin, Y.
\newblock Deep learning in statistical downscaling for deriving high spatial resolution gridded meteorological data: A systematic review.
\newblock \emph{ISPRS Journal of Photogrammetry and Remote Sensing}, 208:\penalty0 14--38, 2024.

\bibitem[Vandal et~al.(2017)Vandal, Kodra, Ganguly, Michaelis, Nemani, and Ganguly]{vandal2017deepsd}
Vandal, T., Kodra, E., Ganguly, S., Michaelis, A., Nemani, R., and Ganguly, A.~R.
\newblock Deepsd: Generating high resolution climate change projections through single image super-resolution.
\newblock In \emph{Proceedings of the 23rd acm sigkdd international conference on knowledge discovery and data mining}, pp.\  1663--1672, 2017.

\bibitem[Wu et~al.(2024)Wu, Chen, Wang, Peng, Sun, and Chen]{wu2024weathergnn}
Wu, B., Chen, W., Wang, W., Peng, B., Sun, L., and Chen, L.
\newblock Weathergnn: Exploiting meteo-and spatial-dependencies for local numerical weather prediction bias-correction.
\newblock In \emph{Proceedings of the International Joint Conference on Artificial Intelligence}, pp.\  2433--2441, 2024.

\bibitem[Wu et~al.(2023)Wu, Zhou, Long, and Wang]{wu2023interpretable}
Wu, H., Zhou, H., Long, M., and Wang, J.
\newblock Interpretable weather forecasting for worldwide stations with a unified deep model.
\newblock \emph{Nature Machine Intelligence}, 5\penalty0 (6):\penalty0 602--611, 2023.

\bibitem[Xiang et~al.(2024)Xiang, Jin, Dong, Bai, Fang, Zhao, Sun, Thambiratnam, Zhang, and Huang]{xiang2024adaf}
Xiang, Y., Jin, W., Dong, H., Bai, M., Fang, Z., Zhao, P., Sun, H., Thambiratnam, K., Zhang, Q., and Huang, X.
\newblock Adaf: An artificial intelligence data assimilation framework for weather forecasting.
\newblock \emph{arXiv preprint arXiv:2411.16807}, 2024.

\bibitem[Xiao et~al.(2024)Xiao, Bai, Xue, Chen, Chen, Chen, Han, and Ouyang]{pmlr-v235-xiao24a}
Xiao, Y., Bai, L., Xue, W., Chen, H., Chen, K., Chen, K., Han, T., and Ouyang, W.
\newblock Towards a self-contained data-driven global weather forecasting framework.
\newblock In \emph{Proceedings of the 41st International Conference on Machine Learning}, volume 235 of \emph{Proceedings of Machine Learning Research}, pp.\  54255--54275. PMLR, 21--27 Jul 2024.

\bibitem[Xie et~al.(2022)Xie, Takikawa, Saito, Litany, Yan, Khan, Tombari, Tompkin, Sitzmann, and Sridhar]{xie2022neural}
Xie, Y., Takikawa, T., Saito, S., Litany, O., Yan, S., Khan, N., Tombari, F., Tompkin, J., Sitzmann, V., and Sridhar, S.
\newblock Neural fields in visual computing and beyond.
\newblock In \emph{Computer Graphics Forum}, volume~41, pp.\  641--676. Wiley Online Library, 2022.

\bibitem[Yang et~al.(2024)Yang, Giezendanner, Civitarese, Jakubik, Schmitt, Chandra, Vila, Hohl, Hill, Watson, et~al.]{yang2024multi}
Yang, Q., Giezendanner, J., Civitarese, D.~S., Jakubik, J., Schmitt, E., Chandra, A., Vila, J., Hohl, D., Hill, C., Watson, C., et~al.
\newblock Multi-modal graph neural networks for localized off-grid weather forecasting.
\newblock \emph{arXiv preprint arXiv:2410.12938}, 2024.

\bibitem[Zhao et~al.(2024)Zhao, Bian, Ni, Jin, Weyn, Fang, Xiang, Dong, Zhang, Sun, et~al.]{zhao2024omg}
Zhao, P., Bian, J., Ni, Z., Jin, W., Weyn, J., Fang, Z., Xiang, S., Dong, H., Zhang, B., Sun, H., et~al.
\newblock Omg-hd: A high-resolution ai weather model for end-to-end forecasts from observations.
\newblock \emph{arXiv preprint arXiv:2412.18239}, 2024.

\bibitem[Zhong et~al.(2024)Zhong, Du, Chen, Wang, and Li]{zhong2024investigating}
Zhong, X., Du, F., Chen, L., Wang, Z., and Li, H.
\newblock Investigating transformer-based models for spatial downscaling and correcting biases of near-surface temperature and wind-speed forecasts.
\newblock \emph{Quarterly Journal of the Royal Meteorological Society}, 150\penalty0 (758):\penalty0 275--289, 2024.

\end{thebibliography}
\bibliographystyle{icml2025}

\newpage
\appendix
\onecolumn

\section{Dataset Description and Preprocessing}\label{sec:Dataset} 
This section presents an overview of the datasets used in this paper - HRRR meteorological analysis data \cite{dowell2022high}, Weather-5K in-situ observational data \cite{han2024weather}, and GEBCO2024 topographical data \cite{mayer2018nippon} - along with their preprocessing methods.

\subsection{NOAA High-Resolution Rapid Refresh \cite{dowell2022high}}
The High-Resolution Rapid Refresh (HRRR) is a weather prediction model developed by the National Oceanic and Atmospheric Administration (NOAA). It provides high-frequency, short-term forecasts with a focus on the United States. Updated hourly, HRRR offers detailed data on various meteorological parameters, such as temperature, wind speed, precipitation, and cloud cover, with a high spatial resolution. This model is particularly useful for applications requiring precise, near-term weather predictions, including aviation, severe weather monitoring, and energy management. The raw dataset is available at \url{https://registry.opendata.aws/noaa-hrrr-pds/}.

In this paper, we use the 0-hour analysis field from the High-Resolution Rapid Refresh (HRRR) model as the input gridded meteorological field. To facilitate spatial alignment with latitude-longitude coordinates and other auxiliary data, we interpolated the original data from irregular grid mapping onto a regular latitude-longitude grid with a resolution of 0.03125$^\circ$ (approximately 3km). The mapped gridded HRRR dataset covers longitudes from 74°W to 121°W and latitudes from 25°N to 47°N. We cropped data from three target subregions as our study areas.
We focus on two near-surface variables: wind speed ($gust$), and 2-meter air temperature ($t_{2m}$). 

\subsection{Weather-5K \cite{han2024weather}}
Weather-5K is a global station forecast dataset proposed by HKUST and Shanghai Artificial Intelligence Laboratory, among other institutions. It covers 5,672 stations worldwide with hourly data over a 10-year period and includes various surface meteorological variables. This dataset integrates and quality-controls data from multiple original observational sources, making it significant for station-based weather forecasting and station-level evaluation of medium-range forecast models. The raw dataset is available at \url{https://hkustconnect-my.sharepoint.com/:u:/g/personal/thanad_connect_ust_hk/EZGm7DP0qstElZwafr_U2YoBk5Ryt9rv7P31OqnUBZUPAA?e=5r0wEo}.

In this study, we selected observation stations within the target study areas and chose wind speed and 2m temperature as the primary variables for an investigation to align with the HRRR data. We used this data as sparse observational supervision for the model and as ground truth labels to validate the effectiveness of our methods. In addition to meteorological variables, the Weather-5K dataset also contains latitude, longitude, and elevation information for observation stations, which we also incorporated as model inputs.

\subsection{GEBCO-2024 \cite{mayer2018nippon}}
The GEBCO-2024 dataset represents the latest global bathymetric and topographic model released by the General Bathymetric Chart of the Oceans (GEBCO). This grid provides continuous, high-resolution elevation data covering Earth's land and ocean floor at 15 arc-second intervals. The dataset integrates various data sources, including ship-based echo-sounding measurements, satellite-derived bathymetry, and land topography from multiple national and regional datasets. The dataset is freely available to users worldwide and is provided in several common GIS formats. It represents a collaborative effort between the International Hydrographic Organization (IHO) and the Intergovernmental Oceanographic Commission (IOC) of UNESCO, continuing GEBCO's tradition of mapping the world's oceans since 1903. The raw dataset is available at \url{https://download.gebco.net/}.

In this study, we downsampled the original resolution data to 0.03125° and cropped the topographic data of the target areas as an auxiliary input for model training. During the inference phase, we interpolated from the original data to obtain 0.015625° according to the target resolution requirements.

\subsection{Preprocessing}
To accommodate the training and testing process of our model, we preprocessed the original data. Specifically, during model input, we normalized the original data using mean and variance. For temperature variables, we standardized the units to Kelvin ($K$). For auxiliary inputs such as topography, latitude/longitude coordinates, and dates, we also applied normalization to the data. We divided the dataset based on time periods. Specifically, data from January 1, 2017, to August 31, 2020, was used as the training set; data from September 1, 2020, to December 31, 2020, as the validation set; and data from January 1, 2021, to December 31, 2021, as the test set.

\begin{table}[]
    \centering
    \begin{tabular}{c|c|c|c}
    \toprule
         Model & Param. Num. & Training Time per Batch (s) & Inference FPS\\\hline
         MLP&75.4K&$\approx 0.5$&$\approx 60$\\
         HyperMLP&1.406M&$\approx 0.5$&$\approx 45$\\
         KANI (Ours)&1.404M&$\approx 1.1$&$\approx 43$\\
         \bottomrule
    \end{tabular}
    \caption{Comparison of Model Parameters and Computational Speed}
    \label{tab:apd_comp}
    \vspace{-20pt} 
\end{table}

\vspace{-5pt} 
\section{Baseline Models}\label{sec:Baseline}

Considering our goal of correcting grid-station bias and performing downscaling under sparse station supervision, we modeled the coordinate-to-meteorological state mapping with gridded meteorological fields as conditions. This approach differs fundamentally from the existing field-to-field post-processing paradigm, thus lacking existing methods for meaningful comparison. Therefore, to demonstrate the innovation and effectiveness of our proposed model more convincingly, we specifically constructed several basic baseline methods for comparison. 

It should be noted that our main objective is to compare how different model structures affect performance, so we \textbf{ensured consistency in input and supervision information across different methods}, although the input data format is also one of our method's innovations (the impact of different input data on performance is discussed in the ablation studies). 

Table \ref{tab:apd_comp} compares the parameter counts and computation speeds between our proposed KANI method and other baseline models. The baseline methods we constructed will be introduced below:

\subsection{Interpolation of Meteorological Field}
The most direct and common method for evaluating gridded meteorological field data at sub-grid observation stations is to interpolate the meteorological field data based on latitude and longitude coordinates. Previous forecasting models based on gridded meteorological fields typically also use interpolation methods when evaluating at observation stations \cite{wu2023interpretable, han2024fengwu}. Therefore, we use interpolation methods as the most basic baseline for comparison. We specifically employed two interpolation methods: linear interpolation and nearest neighbor interpolation, implemented through ${\rm xarray.DataArray.interp()}$ interface.

\subsection{Pure Multilayer Perceptron (MLP)}
Since we need to construct a continuous representation from coordinates to meteorological states, using a pure MLP is one of the most direct model architectures, which has been widely applied in neural radiance fields research, particularly for continuous modeling of 3D scenes and objects, commonly known as NeRF \cite{mildenhall2021nerf}.

In this paper, we designed the MLP according to our task characteristics. Specifically, for the sum of input data embedding and auxiliary input embeddings, we first use a neck FC layer to map from the embedding dimension (64) to the hidden dimension (128), followed by two FC layers with dimension 128, and finally predict the target meteorological states for both grid points and stations through an output FC layer.



\subsection{HyperNetwork with MLP Decoder (HyperMLP)}
To establish a stronger baseline, we constructed a hypernetwork model architecture based on our proposed KANI model, replacing the KAN layers in KANI with traditional MLP layers. Compared to pure MLP and KAN models, the hypernetwork architecture can extract deep semantic features from the input grid meteorological fields and transmit information to the MLP reconstructor through weight generators, enhancing the model's modeling capability and computational complexity and thus increasing model capacity.


\section{More Visualization Results}
\subsection{Bias Correction Results}
In this section, we present additional visualizations of bias correction results to further validate our proposed method's effectiveness.

\begin{figure*}
    \centering
    \includegraphics[width=0.99\linewidth]{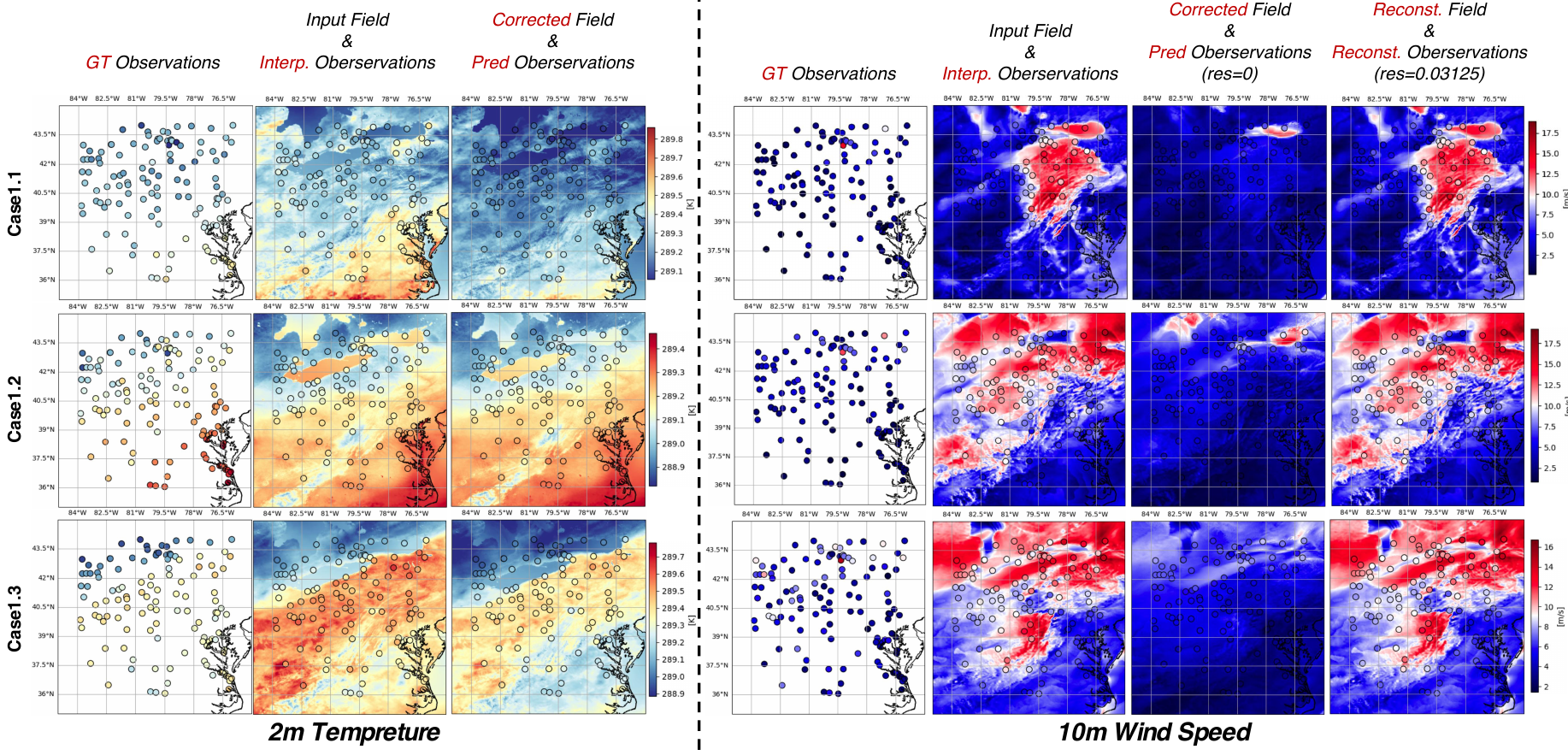}
    \caption{The illustration of proposed KANI for correcting the grid-station systematic bias for the input meteorological field in Region 1.}
    \label{fig:BC}
\end{figure*}

\begin{figure*}
    \centering
    \includegraphics[width=0.99\linewidth]{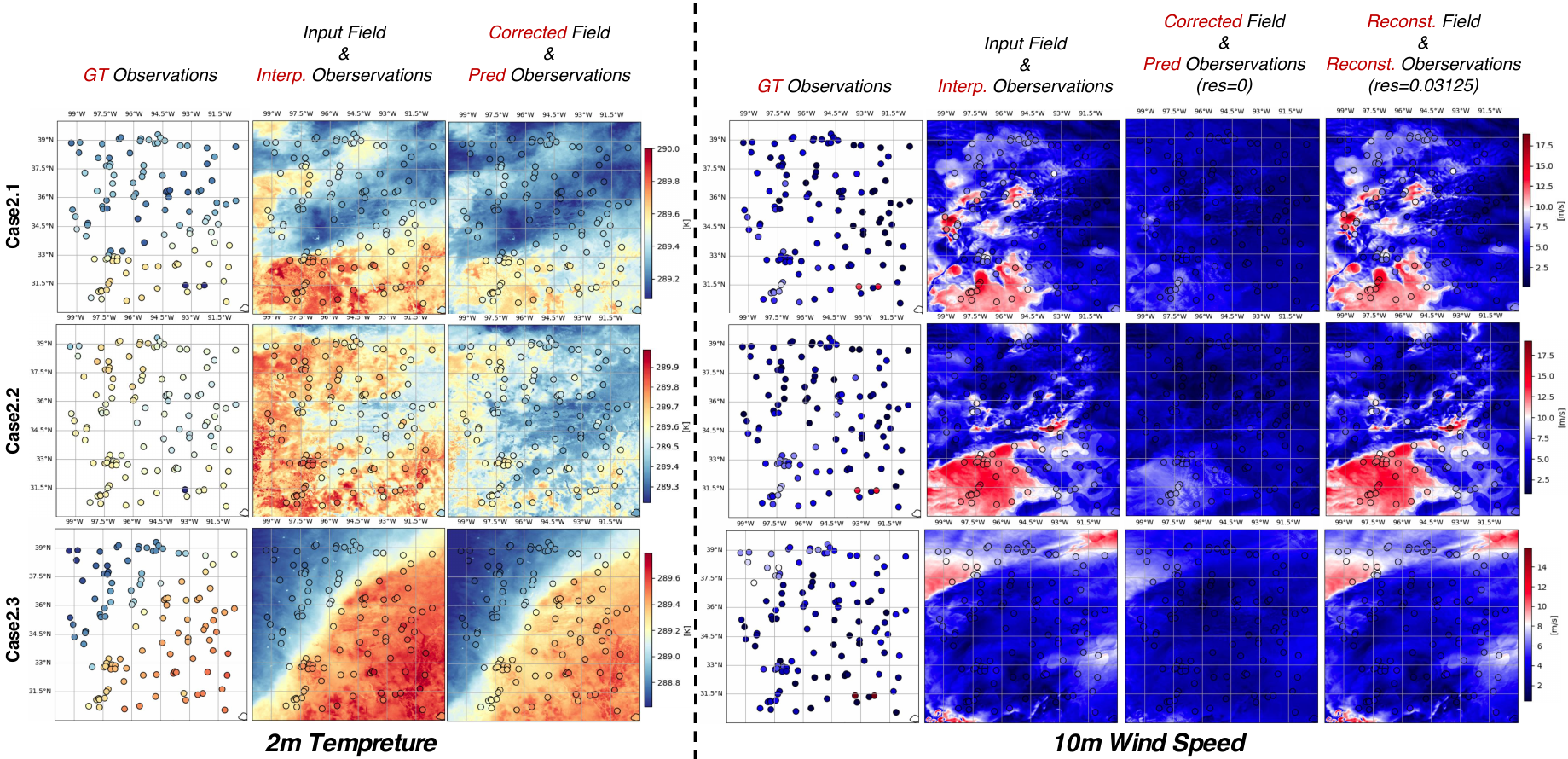}
    \caption{The illustration of proposed KANI for correcting the grid-station systematic bias for the input meteorological field in Region 2.}
    \label{fig:BC}
\end{figure*}

\begin{figure*}
    \centering
    \includegraphics[width=0.99\linewidth]{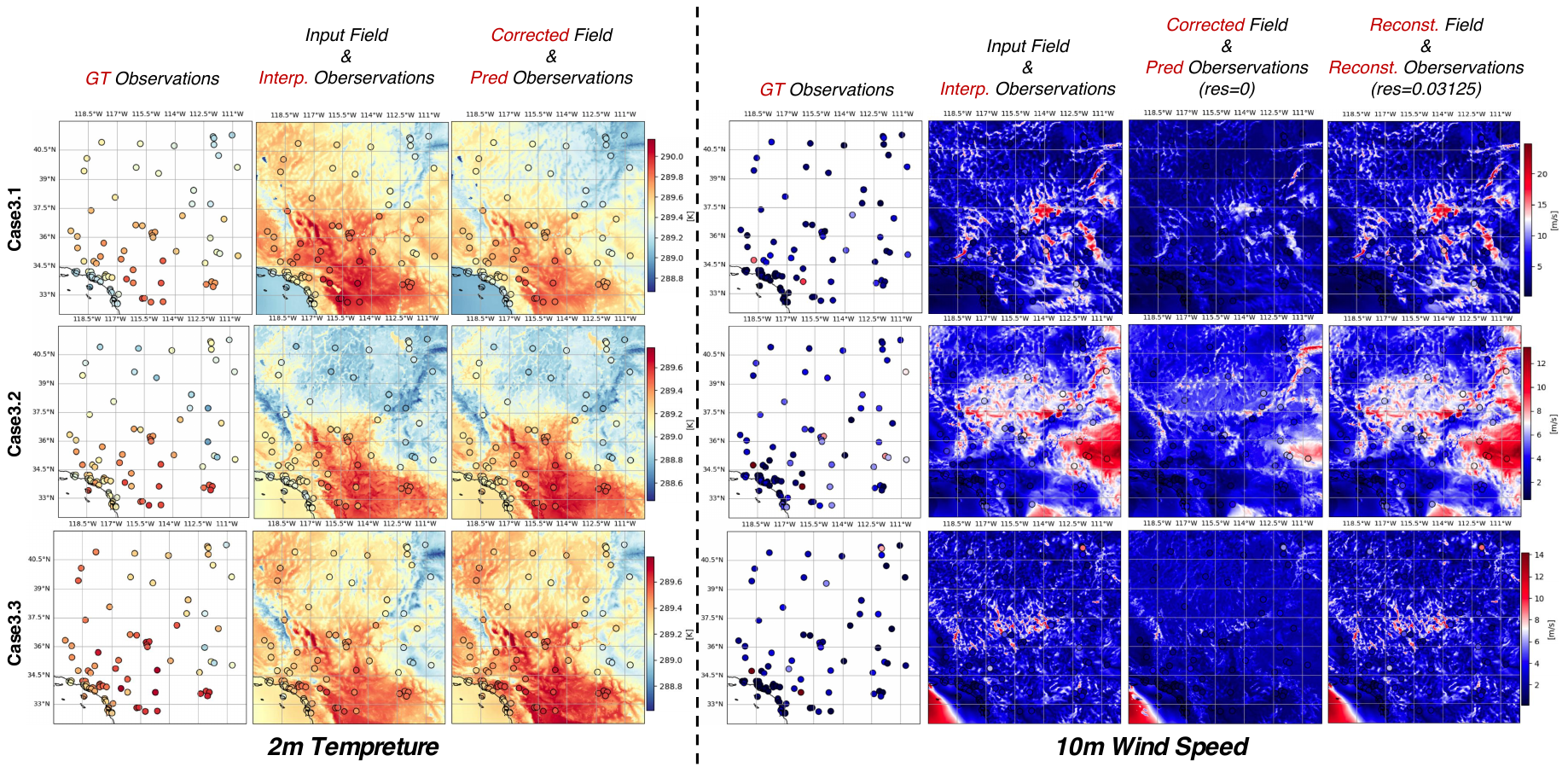}
    \caption{The illustration of proposed KANI for correcting the grid-station systematic bias for the input meteorological field in Region 3.}
    \label{fig:BC}
\end{figure*}


\end{document}